\pdfoutput=1

\documentclass{article}

\usepackage{microtype}
\usepackage{graphicx}
\usepackage{subcaption}
\usepackage{booktabs}

\usepackage{hyperref}

\usepackage[accepted]{icml2026}

\usepackage{amsmath}
\usepackage{amssymb}
\usepackage{mathtools}
\usepackage{amsthm}

\usepackage[capitalize,noabbrev]{cleveref}

\theoremstyle{plain}

\theoremstyle{definition}

\theoremstyle{remark}

\icmltitlerunning{Effects of Structural Reward Shaping on Biophysical Properties in RL-Trained Plasmid Generators}

\begin{document}

\twocolumn[
\icmltitle{\texorpdfstring{Effects of Structural Reward Shaping on Biophysical Properties\\in RL-Trained Plasmid Generators}{Effects of Structural Reward Shaping on Biophysical Properties in RL-Trained Plasmid Generators}}

\begin{icmlauthorlist}
\icmlauthor{McClain Thiel}{ucl}
\icmlauthor{Angus G. Cunningham}{ucl}
\icmlauthor{Chris P. Barnes}{ucl}
\end{icmlauthorlist}

\icmlaffiliation{ucl}{Department of Cell and Developmental Biology, University College London, London, UK}

\icmlcorrespondingauthor{Chris P. Barnes}{christopher.barnes@ucl.ac.uk}

\icmlkeywords{Machine Learning, DNA Language Models, Plasmid Design,
Reinforcement Learning, Group Relative Policy Optimization, Synthetic
Biology}

\vskip 0.3in
]

\printAffiliationsAndNotice{}

\begin{abstract}
We compare the efficacy and distributional effects of supervised
fine-tuning (SFT) and reinforcement learning (RL) post-training for
PlasmidGPT, a foundation model for whole-plasmid generation, using Group
Relative Policy Optimization (GRPO) for the RL model. Using a biologically
motivated reward function encoding functional annotations, length
constraints, and repeat penalties, the RL model achieves a 71.6\%
quality-control pass rate across 8 prompts on 4{,}000 sequences,
compared to 4.3\% for the pretrained baseline and 11.0\% for SFT. A
five-model reward ablation identifies the cassette arrangement
bonus, which rewards correct
promoter$\rightarrow$CDS$\rightarrow$terminator ordering, as the
critical reward component. Rejection-sampling baselines indicate that
the gain is not recovered by sampling up to $K{=}64$ candidates from the
base model (54.5\%), while matching single-sample RL requires $16\times$
more rejection samples from SFT.
Beyond directly optimized features, RL-generated sequences converge
toward real plasmid distributions in 3-mer composition and minimum free
energy density, neither of which is directly optimized by the reward
function. Minimum free energy density independently converges to the
real-plasmid regime under
both SFT and RL despite these being parallel post-training paths. On 29
curated non-Addgene held-out plasmids, RL improves per-token continuation
log-likelihood over the pretrained baseline on all 29 sequences (mean
$\Delta = +0.83$ nats), though SFT achieves higher predictive
log-likelihood ($-7.35$ vs.\ $-8.16$ nats).
\end{abstract}

\section{Introduction}

Plasmids are extrachromosomal DNA sequences that are often found in
bacteria and are capable of replication
independent of a host genome \cite{lederberg1952}. These genetic
elements are ubiquitous in biotechnology, serving as the primary vectors
for protein expression, gene editing, and emerging DNA therapeutics
\cite{plasmid_manufacturing, plasmid_therapeutics}. Despite their
widespread utility, plasmid engineering remains a complex,
high-dimensional optimization problem. Traditional workflows are
cost-intensive and heuristic-driven, often requiring iterative cycles of
manual sequence editing and experimental validation
\cite{plasmid_design_challenges1}. Suboptimal
architectures with incompatible regulatory elements or unstable repeat
regions can lead to metabolic burden, reduced expression efficiency, and
manufacturing bottlenecks \cite{structural_instability,
plasmid_design_challenges2, metabolic_burden}.

Current approaches to plasmid design rely heavily on tacit domain
knowledge and piecemeal assembly of genetic parts \cite{clonefast}.
Designers must simultaneously optimize for competing objectives such as
copy number, transcriptional output, and host viability while navigating
the strict biophysical constraints of DNA folding and context-dependent
regulatory interactions \cite{gene_to_performance}.

Reinforcement learning post-training has proven effective in natural
language processing for improving instruction following and reasoning
\cite{instructgpt, grpo}. We
investigate whether structural reward shaping can similarly improve
generative DNA models, applying Group Relative Policy Optimization (GRPO) to the PlasmidGPT foundation model
\cite{plasmidgpt} for whole-plasmid generation. Our model achieves a
71.6\% quality-control pass rate compared to 4.3\% for the pretrained
baseline across an 8-prompt evaluation on 4{,}000 sequences. Beyond
directly optimized features, generated sequences align with real
plasmids in 3-mer composition. Both
SFT (next-token loss) and RL (GRPO reward shaping starting from Base)
independently converge to plasmid-like 3-mer composition and MFE
density. Across 29 non-Addgene held-out plasmids, RL outperforms the
pretrained baseline on every plasmid for both per-token continuation
log-likelihood and CDS-junction surprisal; within this benchmark, RL's reward
shaping does not degrade next-token prediction relative to Base, though
SFT remains stronger on predictive log-likelihood ($-7.35$ vs.\ $-8.16$
nats).

This study is scoped to engineered \textit{E.~coli} expression vectors
using the Addgene repository as our target distribution. Natural and
environmental plasmids (e.g., PLSDB) are excluded because our reward
function and QC pipeline are designed for engineered constructs with
known functional annotations.

Our contributions are: (i) the first application of GRPO to
whole-plasmid generation, evaluated across 8 diverse prompts on 4{,}000
sequences; (ii) a five-model reward ablation identifying the cassette
bonus as the critical reward component, whose removal drops pass rate
from 66.9\% to 19.8\% at the ablation evaluation temperature
($T=0.95$; Appendix~\ref{sec:ablation_appendix}); and (iii) rejection sampling baselines
demonstrating that RL produces genuine distributional shift rather than
selective filtering (Base reaches 54.5\% at $K{=}64$, while
single-sample RL matches $K{=}16$ SFT rejection sampling;
Table~\ref{tab:baselines}).

Code, trained model weights, and evaluation data are released at
\url{https://github.com/UCL-CSSB/PlasmidRL}.

\section{Background}

\subsection{DNA Language Models}

Foundation models pretrained on large corpora have shown strong transfer
to genomic tasks, including variant prediction \cite{dnabert},
transcription factor binding \cite{hyenadna}, and whole-genome
generation \cite{evo}, although general prokaryotic sequence models are
not post-trained to enforce the strict circular topology, single-origin,
and selection-marker constraints of engineered cloning vectors.

Plasmid DNA has received relatively little attention compared to other
sequence types despite its importance in biomanufacturing and
research. OriGen \cite{origen} introduces a generative model to produce
previously undiscovered origins of replication (ORIs) but does not model
whole sequences. PlasmidGPT \cite{plasmidgpt} uses modern language
modeling techniques to develop a generative model for whole plasmid
sequences, and later work expands on this by synthesizing whole plasmids
generated using a fine-tuned version of the PlasmidGPT model
\cite{angus}.

Other recent sequence-design methods focus on shorter regulatory loci:
GENERator \cite{generator} is a 1.2B-parameter long-context
autoregressive genomic foundation model applied to cis-regulatory
element design, and TACO \cite{regulatory_dna_rl} uses reinforcement
learning to fine-tune HyenaDNA for promoter and enhancer design
($<$1\,kb). Neither targets whole multi-gene plasmid
constructs of several kilobases. Our work extends RL post-training
to whole-plasmid generation, where the model must coordinate multiple
functional components simultaneously across several kilobases of
sequence.

\subsection{Plasmid Design}

Lab-designed plasmids are short circular DNA molecules (typically 2--15\,kb)
that must contain multiple functional components arranged in precise
configurations. At minimum, a viable plasmid requires: (i) an origin of
replication to enable autonomous replication, (ii) a selection
marker (e.g., antibiotic resistance gene) for identifying successfully
transformed cells, and (iii) a cloning site where genes of interest can
be inserted.

The search space of valid plasmids is massive due to combinatorial
explosion across components \cite{combinatorial_explosion}. Each
component class can contain hundreds to thousands of variants, and
additional regulatory elements (promoters, terminators, enhancers) and
reporters may be required depending on the application. Multiple
instances of some components may be necessary, and their ordering and
spacing significantly affect function. Beyond sheer
combinatorics, designers must navigate complex biological constraints
such as compatibility requirements (specific ORIs only function in
certain hosts), physical stability issues (repeat regions can fold and
bind to each other), and other design challenges \cite{plasmid_design_challenges1, plasmid_design_challenges2}.

\section{Methods}

\begin{figure*}[!t]
\vskip 0.2in
\begin{center}
\includegraphics[width=\textwidth]{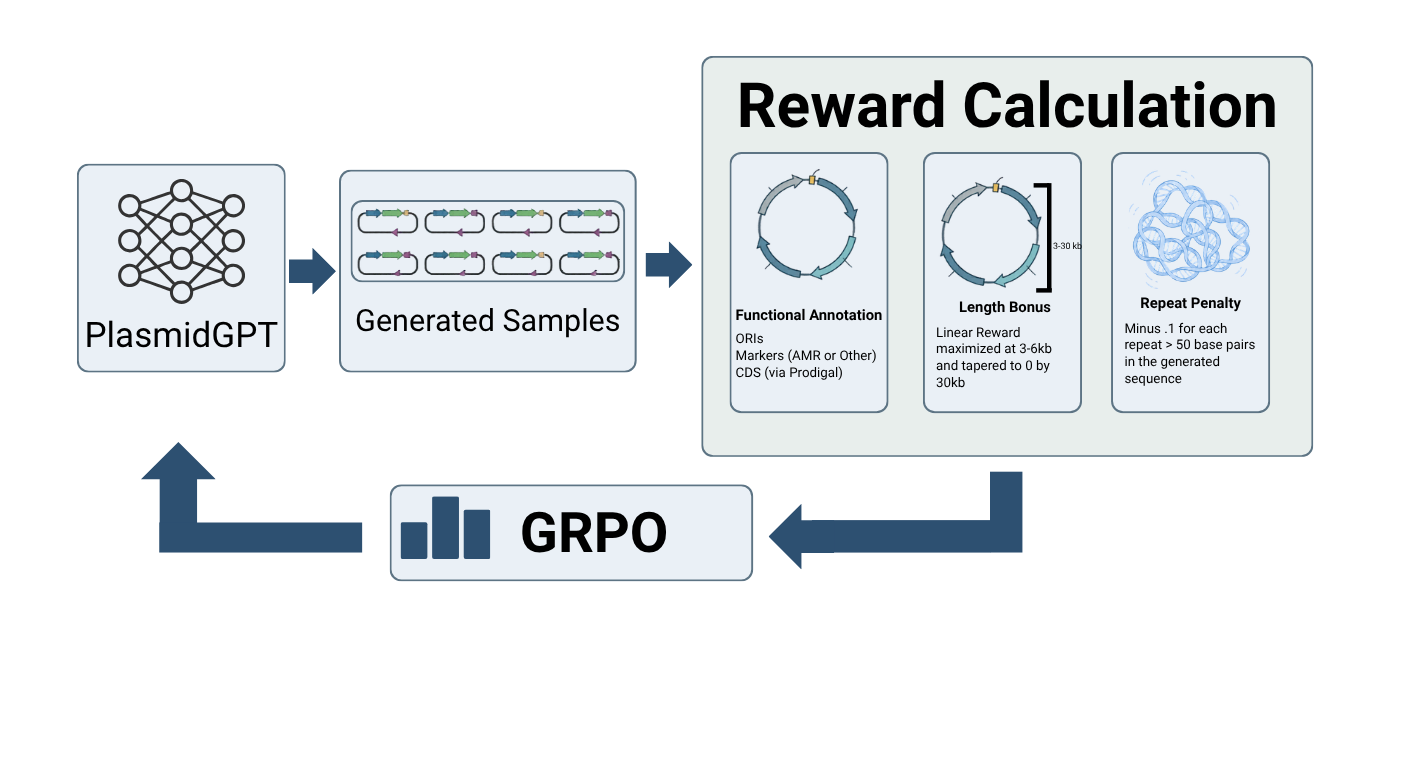}
\caption{Plasmid-RL training pipeline. GRPO is applied to PlasmidGPT using a reward function encoding functional annotations, length constraints (extra bonus maximized at 3--6\,kb and tapering linearly to zero at 2\,kb and 30\,kb), and repeat content.}
\label{fig:methods_overview}
\end{center}
\vskip -0.2in
\end{figure*}

We compare GRPO against two baselines: the pretrained PlasmidGPT model
and a supervised fine-tuned (SFT) baseline from prior lab work
\cite{angus}. That work introduced the curated SFT corpus and
bioinformatics QC pipeline, including limited wet-lab validation of
sequences passing the screen; we reuse the same QC criteria here and
train an RL model optimized for sequence-level structural rewards
(Figure~\ref{fig:methods_overview}).

\subsection{Supervised Fine-Tuning}

The SFT baseline follows prior work from our lab \cite{angus}. Briefly,
supervised fine-tuning was performed on a curated corpus of
\textit{E.~coli} plasmid sequences assembled from PlasmidScope and
Addgene \cite{plasmidscope, addgene}. After deduplication and quality filtering, approximately 15k
circular plasmids ($\leq$30\,kb) were retained, excluding linear
entries, fragments, and incomplete records. Sequences were tokenized
using the original PlasmidGPT byte-pair encoding (BPE) DNA tokenizer,
where tokens average 20--60 nucleotides; consequently, the 256-token
generation limit used during RL corresponds to approximately
5--15\,kb of sequence. The pretrained
PlasmidGPT model was fine-tuned using an autoregressive next-token
prediction objective with gradient accumulation and learning-rate warmup
over three epochs.

\subsection{Reinforcement Learning with GRPO}

We implement a configurable reinforcement learning pipeline for plasmid
design that uses Group Relative Policy Optimization (GRPO) \cite{grpo}
with a domain-specific reward function described below. At each training
iteration, the model generates a batch of candidate plasmids via
autoregressive rollouts conditioned on short nucleotide prompts. Prompts
are either stochastic (4--25 bp random seeds, excluding ``ATG'', used to
promote rollout diversity) or structured (partial ``cassette'' seeds
encoding canonical marker genes such as antibiotic-resistance or
fluorescent reporters). Each candidate sequence is evaluated via our
reward function, which captures structural plausibility, cassette
organization, repeat content, and other biologically motivated
constraints. GRPO is then applied to update the model parameters using
these sequence-level rewards, enabling the policy to progressively shift
toward generating plasmids with higher predicted validity. GRPO training
hyperparameters and the length-bonus scale were selected via an Optuna
sweep maximizing mean training reward on held-in training prompts (with
the 8 evaluation prompts excluded), whereas SFT retains the fixed
hyperparameters of \citet{angus}; sampling temperatures follow a
per-model temperature sweep ($n{=}200$ per setting; Appendix~\ref{sec:sampling_params}).

\subsection{Reward Function Design}

The reward function scores each generated plasmid according to its
structural plausibility and expected stability. It is composed of three
conceptual components:

\textbf{Functional annotation scoring:} Lightweight rollout annotations
identify ORIs, promoters, terminators, and selectable markers via
PlasmidKit (\url{https://github.com/McClain-Thiel/plasmid-kit}), which
combines Aho--Corasick seed matching and Edlib alignment against a
curated plasmid-part library, and then score each feature class according
to SME-defined count targets (e.g., exactly one origin of replication and
at least one selectable marker). CDS regions are identified using
Pyrodigal \cite{prodigal, pyrodigal}, a gene prediction tool that
detects coding sequences based on statistical patterns rather than
homology search. To encourage coherent gene cassettes, this component
also includes a location-aware bonus for promoter $\rightarrow$ CDS
$\rightarrow$ terminator arrangements that appear in the correct order
and within a reasonable proximity window.

\textbf{Length prior:} A length prior favors plasmid sizes within typical
experimental ranges preferred for plasmid construction. The length factor
is 1 plus an extra bonus inside the acceptable range: sequences between
3\,kb and 6\,kb receive the full extra length bonus, and this extra
bonus tapers to zero at the acceptable-range bounds (2\,kb and
30\,kb); sequences outside [2\,kb, 30\,kb] receive length factor 0
(under the 256-token rollout cap, $\approx$5--15\,kb, sequences rarely
enter the upper 15--30\,kb taper).

\textbf{Repeat penalty:} A repeat penalty down-weights sequences
containing long exact repeats that are associated with instability or
recombination, subtracting 0.1 reward for each repeat of length 50 bp
or greater.

These terms are combined into a single scalar in $[0,1]$, yielding a
fast and interpretable proxy for ``plasmid-likeness'' and validity during
reinforcement learning. Algorithm~\ref{alg:reward} gives the full
scoring pipeline and Algorithm~\ref{alg:cassette} the cassette-bonus
subroutine; component weights and count thresholds are listed in
Appendix~\ref{sec:reward_config}.

\section{Experiments}

\subsection{Plasmid Quality Control and Uniqueness}

We evaluate model variants across 8 diverse prompts spanning 3--917\,bp:
a minimal start codon (ATG, 3\,bp), random short seeds (10\,bp,
25\,bp), replication-origin prefixes (100\,bp from pUC19 and
pACYC184/p15A), a resistance cassette (KanR, 300\,bp), a viral
regulatory element (CMV enhancer, 300\,bp), and a full GFP expression
cassette (917\,bp). For each prompt, 500 sequences were generated for a total of
4{,}000 sequences per model. Prompt strings were excluded from the RL prompt distribution;
the underlying biological motifs may still occur in the pretraining
or SFT corpora, so evaluation reflects generalization at the prompt
level rather than the motif level. Notably, ATG is explicitly
excluded from the training prompt distribution (see
Appendix~\ref{sec:rl_hparams}), so its inclusion here tests
out-of-distribution behavior.

\subsubsection{Validity Assessment}
\label{sec:validity}

In silico plasmid validity was assessed using the two-stage bioinformatics
quality-control pipeline introduced in our prior lab work \cite{angus},
which requires exactly one origin of replication ($\geq$99\% identity and
subject coverage via BLAST \cite{blast} against a curated OriDB reference,
with no secondary hits $\geq$85\% identity and $\geq$80\% coverage), one or
two antimicrobial resistance genes called by AMRFinderPlus \cite{amrfinderplus}
with 100\% identity and coverage, and no circular exact direct or inverted
repeats $\geq$50\,bp via suffix-array detection. While the RL reward uses
PlasmidKit to score fast proxies for these same three constraints during
rollouts, strict QC relies on independent alignment/resistance callers and
circular repeat checking. This pipeline has been evaluated with limited
wet-lab validation of screen-passing sequences \cite{angus}. While we do
not perform wet-lab validation in this work, our focus is on
establishing that RL post-training can successfully navigate the plasmid
design space in silico, laying groundwork for future conditional
generation systems where user-specified designs can be experimentally
validated.

\subsubsection{Uniqueness Assessment}
\label{sec:novelty}

To assess whether generated plasmids represent genuinely new designs
rather than minor variants of existing constructs, we query random
unfiltered subsamples of generated plasmids against NCBI via the BLASTn
API (\citealt{blast}; subsample sizes reflect API-completed queries: Base
$n{=}22$, SFT $n{=}28$, RL $n{=}30$). Each generated plasmid is
assigned to one of three categories based on identity and query-coverage
thresholds: sequences with $\geq$99\% identity and $\geq$95\% coverage
are classified as \textbf{Exists}; those with $\geq$95\% identity and
$\geq$80\% coverage are \textbf{Similar}; and all others are classified
as \textbf{Novel}.

This categorization follows large-scale plasmid curation efforts such
as PLSDB, which use similar thresholds to de-duplicate plasmids before
adding sequences to a dataset
\cite{PLSDB}.

\subsubsection{Diversity Assessment}

To monitor potential model collapse, we measure the diversity of many
samples from the same prompt. Standard text-diversity metrics
(e.g., Self-BLEU) are not well-calibrated for biological interpretation
on DNA sequences, so we use the mean pairwise Jaccard distance of the
21-mers of each sequence.
Diversity of a group of rollouts is calculated as follows:

\[D = 1 - \frac{1}{\binom{n}{2}} \sum_{i=1}^{n} \sum_{j=i+1}^{n} J(S_i, S_j)\]

where $J(S_i, S_j)$ is the Jaccard similarity between forward-strand
MinHash sketches ($k{=}21$, sketch size $1{,}000$) \cite{minhash} of
sequences $i$ and $j$, and $n$ is the number of sequences in the group.

The diversity metric (pairwise Jaccard distance) serves primarily as a
model collapse detector rather than a biological validity measure;
$D \approx 1$ indicates near-maximal pairwise dissimilarity, as in
random or otherwise unconstrained sequence generation, while
$D \approx 0$ indicates collapse onto near-identical outputs. We
report numerical values in \S\ref{sec:results}.

\subsubsection{Results}
\label{sec:results}

\begin{table}[t]
\caption{Quality metrics across model variants, all evaluated on 4{,}000
sequences across the same 8 prompts using the QC pipeline of \citet{angus}
(see \S\ref{sec:validity}). Diversity is the mean
pairwise 21-mer Jaccard distance, computed as a pooled distance over
all 4{,}000 sequences (including cross-prompt pairs); RL's mean
within-prompt diversity across the 8 prompts is $0.365$ (median $0.13$;
Table~\ref{tab:per_prompt}).
The Addgene row reports the 500-plasmid reference panel
(Appendix~\ref{sec:reference_plasmids}); QC is not applicable for real
plasmids.}
\label{tab:qc_results}
\vskip 0.15in
\begin{center}
\begin{small}
\begin{sc}
\begin{tabular}{lcc}
\toprule
Model & QC Pass Rate & Diversity \\
\midrule
Addgene ($n{=}500$) & --- & 0.848 \\
Base  & 4.3\%  & 0.960 \\
SFT & 11.0\% & 0.921 \\
RL & 71.6\% & 0.596 \\
\bottomrule
\end{tabular}
\end{sc}
\end{small}
\end{center}
\vskip -0.1in
\end{table}

Reinforcement learning substantially increases the probability of
generating plasmids that pass our bioinformatics quality control (QC)
pipeline, while supervised fine-tuning provides a more modest
improvement.
Table~\ref{tab:qc_results} summarizes headline QC pass rate and
diversity across models.

Aggregated across 8 prompts, the overall QC pass rate
rises from 4.3\% with the base model to 11.0\% with SFT and
71.6\% with RL. Marginal single-criterion pass rates
(single ORI / 1--2 AMR genes / repeat $<$50\,bp) explain these gaps:
Base achieves 17.1\% / 16.2\% / 86.0\% (failing mostly on short
unannotated fragments), SFT reaches 39.0\% / 46.6\% / 57.1\% (frequently
failing multi-ORI calls, 26.3\%, and repeats $\geq$50\,bp, 42.9\%),
and RL reaches 92.1\% / 93.4\% / 77.1\%.
These are parallel post-training baselines: SFT
provides a $\sim$2.6$\times$ improvement over Base, while RL provides
a $\sim$17$\times$ improvement over Base under the same 8-prompt
evaluation.
Per-prompt RL pass rates range from 14.4\% (pUC19 ORI prefix) to 96.6\%
(CMV enhancer); full per-prompt results (including Base and SFT) are in
Appendix~\ref{sec:per_prompt_appendix}. On both 100\,bp ORI-fragment
prompts, RL generates a full-length ColE1 origin downstream (single ORI
in 482/500 pUC19 and 494/500 p15A rollouts); because the pUC19 prompt is
itself a 100\,bp ColE1 substring, appending a full ColE1 creates a
100\,bp duplication that triggers the $\geq$50\,bp repeat filter in
410/500 (82.0\%) rollouts, whereas the heterologous p15A prefix avoids
exact repeats (82.0\% pass).

Pooled across all 4{,}000 sequences, RL diversity is 0.596 compared to
0.960 (near-maximal) for Base, where high diversity reflects
unconstrained, mostly invalid generation, 0.921 for SFT (0.783 among
strict-QC-passing SFT outputs), and 0.848 over the 500-plasmid
Addgene reference panel (Appendix~\ref{sec:reference_plasmids}).
Within-prompt
evaluation (Table~\ref{tab:per_prompt}) reveals a bimodal pattern masked
by pooling: RL's mean within-prompt diversity is 0.365 (median 0.13),
with strong convergence ($D \in [0.08, 0.13]$) on 5 of 8 prompts
(ATG, random 10/25\,bp, p15A ORI, GFP cassette) alongside high diversity
($D \in [0.71, 0.86]$) on KanR, CMV, and pUC19.

Applying the BLASTn novelty assessment from Section~\ref{sec:novelty}
to random unfiltered subsamples (Table~\ref{tab:novelty}), RL yields
0/30 exact matches to known NCBI plasmids (95\% Clopper--Pearson CI
$[0.0\%, 11.6\%]$), with 33.3\% Similar and 66.7\% Novel. Base and
SFT samples are predominantly Novel (90.9\% and 96.4\%), but because
most unfiltered Base and SFT outputs fail QC (with many truncated below
query-coverage thresholds), this reflects unconstrained, low-validity
generation rather than designed novelty. The shift from 3.6--9.1\%
Similar in Base/SFT to 33.3\% Similar in RL is consistent with RL
concentrating probability mass on \textit{E.~coli} expression-vector
backbones without copying single plasmids verbatim in the evaluated
subsample.

\begin{table}[t]
\caption{BLASTn novelty assessment on random unfiltered subsamples
(Base $n{=}22$, SFT $n{=}28$, RL $n{=}30$; 95\% Clopper--Pearson
confidence intervals in brackets). \textbf{Exists}: $\geq$99\%
identity, $\geq$95\% coverage. \textbf{Similar}: $\geq$95\% identity,
$\geq$80\% coverage. \textbf{Novel}: otherwise.}
\label{tab:novelty}
\vskip 0.15in
\begin{center}
\begin{small}
\begin{sc}
\resizebox{\columnwidth}{!}{%
\begin{tabular}{lccc}
\toprule
Model & Exists & Similar & Novel \\
\midrule
Base & 0.0\% {\scriptsize[0.0, 15.4]} & 9.1\% {\scriptsize[1.1, 29.2]}  & 90.9\% {\scriptsize[70.8, 98.9]} \\
SFT  & 0.0\% {\scriptsize[0.0, 12.3]} & 3.6\% {\scriptsize[0.1, 18.3]}  & 96.4\% {\scriptsize[81.7, 99.9]} \\
RL   & 0.0\% {\scriptsize[0.0, 11.6]} & 33.3\% {\scriptsize[17.3, 52.8]} & 66.7\% {\scriptsize[47.2, 82.7]} \\
\bottomrule
\end{tabular}
}
\end{sc}
\end{small}
\end{center}
\vskip -0.1in
\end{table}

\subsection{Reward Ablation Study}

To identify which reward components drive the observed improvements, we
train five ablation models by selectively disabling individual reward
terms (Appendix~\ref{sec:ablation_appendix}). Removing the cassette
arrangement bonus is the most damaging single change, dropping pass
rate from 66.9\% to 19.8\% at $T{=}0.95$ (and from 78.3\% to 44.5\% at
$T{=}1.15$). At $T{=}0.95$, removing the repeat penalty (72.2\%) or
length prior (71.4\%) increases raw pass rate over the full reward by
collapsing onto lower-diversity sequences (0.394 and 0.447 vs.\ 0.578)
with weaker MFE stability, whereas at $T{=}1.15$ the full reward leads
all variants (78.3\% vs.\ 75.2\% and 72.2\%; Figure~\ref{fig:qc_breakdown}).

\subsection{Rejection Sampling Baselines}

To confirm that RL produces genuine distributional shift rather than
simply filtering low-quality sequences, we compare against rejection
sampling at varying $K$ (Table~\ref{tab:baselines}). Within each trial,
$K$ candidates are drawn without replacement from a 1{,}250-sequence pool
per (model, prompt), resampled across $M{=}50$ bootstrap trials per
prompt ($M{\times}8{=}400$ total trials per cell). Temperatures follow a
7-point sweep ($T \in [0.3, 1.3]$, $n{=}200$ per cell), where Base and
SFT peaked at $T{=}1.0$ and RL peaked at $T{=}1.15$; we also include RL at
$T{=}1.0$ for a temperature-matched comparison.

\begin{table}[t]
\caption{Rejection sampling success rate at varying $K$: fraction of
$M{=}50$ trials in which $\geq$1 of $K$ samples passes QC. Base
($T{=}1.0$), SFT ($T{=}1.0$), and RL ($T{=}1.15$) are evaluated at their
sweep-optimal temperatures from the 1{,}250-sequence-per-prompt pools; RL
($T{=}1.0$) reports exact hypergeometric expectation on the 8-prompt
$T{=}1.0$ set ($M{=}50$ bootstrap: 73.0\% / 91.0\% / 98.0\% / 100.0\%).
Because the rejection pools are freshly resampled, $K{=}1$ differs
slightly from Table~\ref{tab:qc_results} for SFT (9.8\% vs.\ 11.0\%)
and for RL at $T{=}1.15$ (76.8\% vs.\ 71.6\% at $T{=}1.0$).}
\label{tab:baselines}
\vskip 0.15in
\begin{center}
\begin{small}
\begin{sc}
\begin{tabular}{lcccc}
\toprule
& $K=1$ & $K=4$ & $K=16$ & $K=64$ \\
\midrule
Base ($T{=}1.0$) & 4.3\%  & 14.5\% & 38.8\% & 54.5\% \\
SFT ($T{=}1.0$)  & 9.8\%  & 36.3\% & 76.3\% & 99.3\% \\
RL ($T{=}1.0$)   & 71.6\% & 92.7\% & 99.0\% & 100.0\% \\
RL ($T{=}1.15$)  & 76.8\% & 95.0\% & 99.0\% & 100.0\% \\
\bottomrule
\end{tabular}
\end{sc}
\end{small}
\end{center}
\vskip -0.1in
\end{table}

Across practical low-to-moderate sampling budgets, the success-rate gap
between RL and Base/SFT is large. RL at $K{=}1$ ($76.8$\% at $T{=}1.15$,
or $71.6$\% at matched $T{=}1.0$) roughly matches SFT at $K{=}16$
($76.3$\%), and RL reaches near-saturation by $K{=}16$ ($99.0$\% at both
temperatures). Base never closes the gap on this protocol: even at
$K{=}64$ Base reaches only $54.5$\%. The $K{=}1$ column recapitulates the
model ordering of Table~\ref{tab:qc_results}; the
$K{\in}\{4,16,64\}$ columns quantify how much rejection sampling can
recover in each model. SFT's near-perfect saturation at $K{=}64$
shows that QC filtering can recover rare valid plasmids even when most
candidates fail. RL's saturation at $K{=}16$ shows that reward-shaped
post-training makes valid plasmids much more common under direct
sampling.

\subsection{Distribution Comparison}
\label{sec:dist_comparison}

We compute biophysical summary statistics from generated and reference
sequences to assess distributional alignment beyond the QC pipeline's
binary pass/fail. The reference panel consists of 500 engineered
plasmids from Addgene used for protein expression, genome editing, and
other applications (see Appendix~\ref{sec:reference_plasmids} for the
selection procedure). Statistics include sequence length, GC content,
Jensen--Shannon divergence \cite{jsd} of the 3-mer composition, and
minimum free energy density (ViennaRNA; \citealt{viennarna}).

\begin{figure*}[!t]
\vskip 0.1in
\begin{center}
\includegraphics[width=0.85\textwidth]{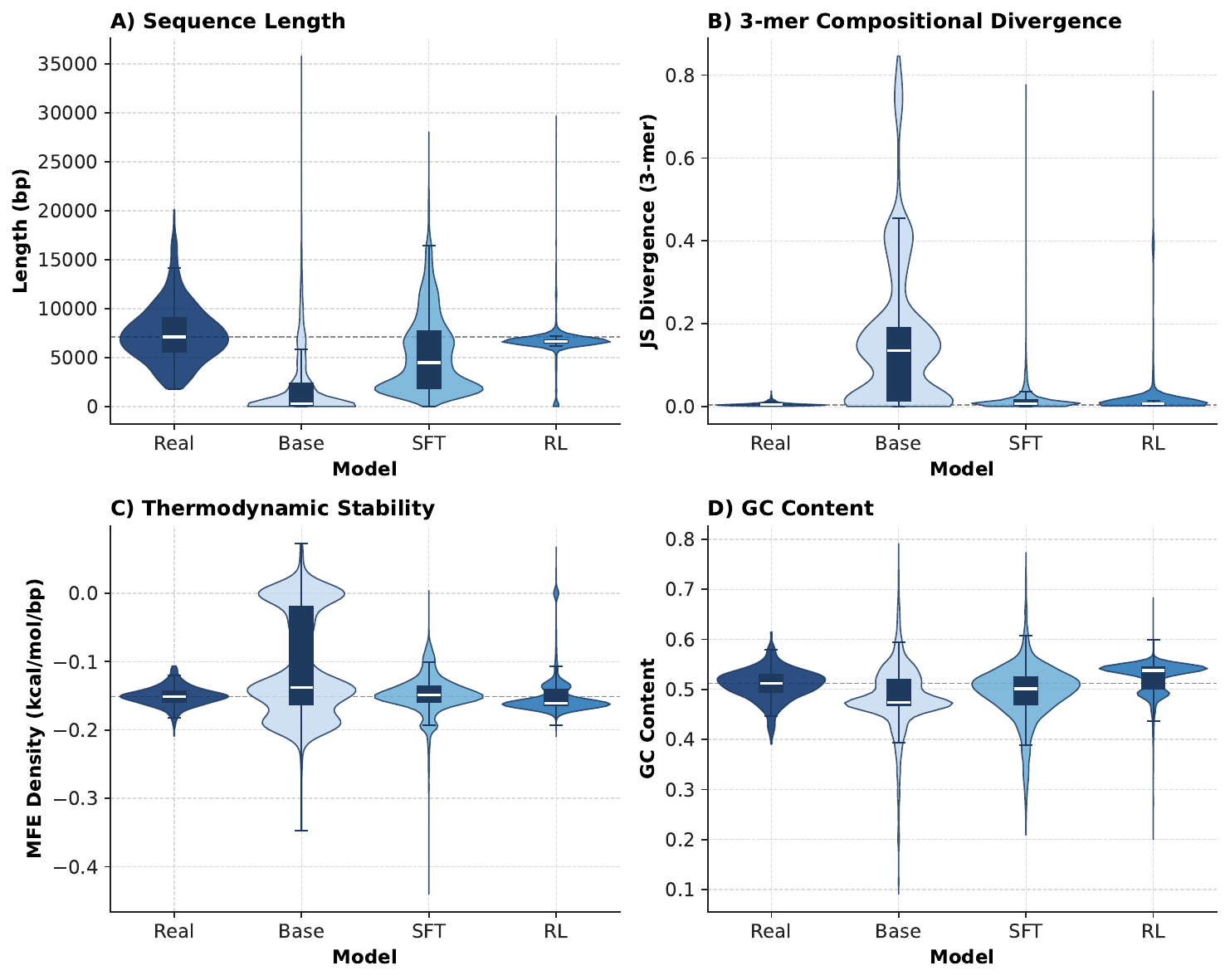}
\caption{Distributional alignment across directly optimized, reward-correlated, and indirectly shaped biophysical metrics.}
\label{fig:distributions}
\end{center}
\vskip -0.2in
\end{figure*}

Figure~\ref{fig:distributions} shows that RL-generated sequences
closely match real plasmids across the displayed metrics, even those not
explicitly rewarded. Quantitative values are in Table~\ref{tab:ks_stats}.
GC content and 3-mer composition both shift toward the real-plasmid
distribution, with SFT and RL reaching comparable 3-mer
divergence despite independent training objectives. RL mean sequence
length (6{,}517\,bp) is closest to the Addgene reference (7{,}469\,bp),
while Base skews short due to its unconstrained generation.

We categorize these distributional effects into three tiers based on
their relationship to the reward function:

\textbf{Directly optimized:} Sequence length is explicitly encoded by
the reward function. RL mean length is closest to the Addgene
reference panel, while Base skews far short and SFT falls between the
two (Table~\ref{tab:ks_stats}).

\textbf{Reward-correlated:} GC content is not directly optimized, but
functional elements selected by the reward (ORIs, coding sequences,
resistance markers) carry GC content characteristic of
\textit{E.~coli} plasmids. RL GC (0.524) slightly overshoots the
real-plasmid value (0.510), suggesting that selection pressure on
GC-rich functional elements pulls composition above the natural
distribution rather than tracking it precisely.

\textbf{Indirectly shaped:} 3-mer composition and minimum free energy
(MFE) density are not named by the reward function, but both shift
sharply in SFT and RL relative to Base. JSD against the reference
3-mer distribution drops in both fine-tuning paths to near the
real-plasmid value (Table~\ref{tab:ks_stats}). MFE density behaves
similarly: SFT ($-0.148$) and RL ($-0.149$) both land within noise of
the real-plasmid reference ($-0.151$), while Base sits far above
($-0.105$). This convergence from two independent post-training
objectives indicates that selecting intact \textit{E.~coli} vector
backbones and Pyrodigal-scored prokaryotic coding sequences (alongside
length constraints) carries canonical local $k$-mer and secondary-structure
statistics without explicitly optimizing either, and that RL does not
achieve high QC scores by producing physically unusual sequences.

\subsection{Held-Out Continuation}
\label{sec:held_out}

To evaluate how RL training affects sequence-level predictive
performance, we measure teacher-forced log-likelihood on real plasmid
continuations. For each sequence, nucleotide windows are sliced with a
400\,bp prefix, 100\,bp target, and 300\,bp stride, and we compute the
per-plasmid mean per-BPE-token log-probability over the target window
(because PlasmidGPT BPE tokens span 20--60\,nt, per-token scores reflect
multi-nucleotide tokens and vary with regional token length).
Table~\ref{tab:continuation} reports the average per-plasmid
log-probability for Base, SFT, and RL on this benchmark.

\begin{table}[t]
\caption{Average per-plasmid, per-BPE-token continuation log-probability
on 29 held-out non-Addgene plasmids (NCBI shuttle vectors, SEVA
backbones, and vectors from \textit{Corynebacterium},
\textit{Mycobacterium}, and \textit{Lactobacillus}). RL improves over
Base on every plasmid (29/29), while SFT provides the strongest
predictive baseline.}
\label{tab:continuation}
\vskip 0.15in
\begin{center}
\begin{small}
\begin{sc}
\begin{tabular}{lcc}
\toprule
Model & Mean LogProb & Error (SEM) \\
\midrule
Base & $-8.99$ & $0.23$ \\
SFT  & $-7.35$ & $0.18$ \\
RL   & $-8.16$ & $0.20$ \\
\bottomrule
\end{tabular}
\end{sc}
\end{small}
\end{center}
\vskip -0.1in
\end{table}

\begin{figure}[t]
\vskip 0.1in
\begin{center}
\includegraphics[width=\columnwidth]{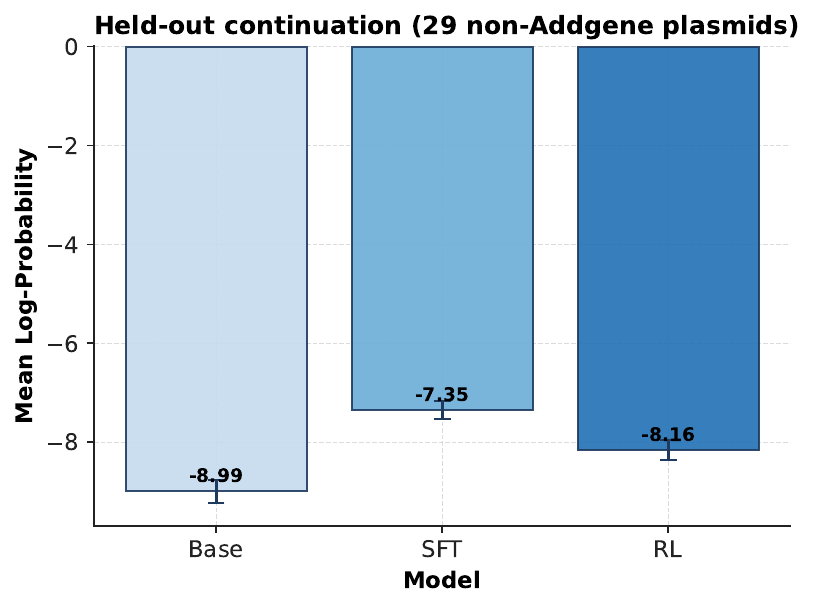}
\caption{Per-plasmid log-probability on 29 held-out non-Addgene
plasmids. RL outperforms Base on every sequence (29/29), with
consistent improvement across diverse plasmid origins. SFT shown for
context.}
\label{fig:completion}
\end{center}
\vskip -0.1in
\end{figure}

The RL model shows a robust improvement in mean log-probability
compared to the base model (mean $\Delta = +0.83$ nats; paired
$t$-test $t=+7.61$, $p=2.75\times10^{-8}$; Wilcoxon signed-rank
$p=3.73\times10^{-9}$; Cohen's $d=+1.41$, large effect; RL beats Base
on 29/29 plasmids). The corresponding standard errors are 0.23 (Base),
0.18 (SFT), and 0.20 (RL). The same RL-over-Base pattern holds for
CDS-junction log-probability (mean $\Delta = +2.40$ nats, $p=2.98\times
10^{-14}$, $d=+2.62$, 29/29). Within this scope we observe no
alignment tax relative to Base. Figure~\ref{fig:completion} shows that
this improvement is consistent across all 29 held-out plasmids. Evaluated
on this same held-out set, SFT achieves higher continuation
log-probability ($-7.35 \pm 0.18$) than RL (mean
$\Delta_{\text{SFT}-\text{RL}} = +0.81$ nats, $p=1.35\times 10^{-6}$,
25/29). SFT is trained directly on the Addgene corpus and provides a
strong predictive baseline that RL does not exceed, while RL strictly
improves over Base on all 29 plasmids ($\Delta = +0.83$ nats, 95\% CI
$[+0.61, +1.05]$). Normalizing total target log-probability per nucleotide
($100$\,bp per window) weights short-token non-Addgene cargo stretches
proportionally to token count and yields near-parity between RL
($-1.02 \pm 0.05$\,nats/nt) and Base ($-1.00 \pm 0.05$\,nats/nt; mean
$\Delta = -0.02$\,nats/nt, Wilcoxon $p = 0.15$; SFT $-0.94 \pm 0.05$\,nats/nt),
indicating that the per-token continuation gain concentrates in
multi-nucleotide conserved motifs with negligible per-base alignment tax
elsewhere.

\subsection{CDS-Junction Log-Probability Analysis}

For each held-out real plasmid, we identify CDS starts via Pyrodigal
\cite{prodigal, pyrodigal} in metagenomic (\texttt{meta}) mode and
compute each model's mean per-BPE-token log-probability over a 100\,bp
($\pm 50$\,bp) nucleotide window centered on each CDS start. We refer to
this evaluation as CDS-junction log-probability (equivalently, negative
surprisal). Whereas the training reward runs Pyrodigal on
\emph{generated} sequences, this evaluation scores held-out \emph{real}
non-Addgene plasmids under teacher forcing.

\begin{figure}[t]
\vskip 0.1in
\begin{center}
\includegraphics[width=\columnwidth]{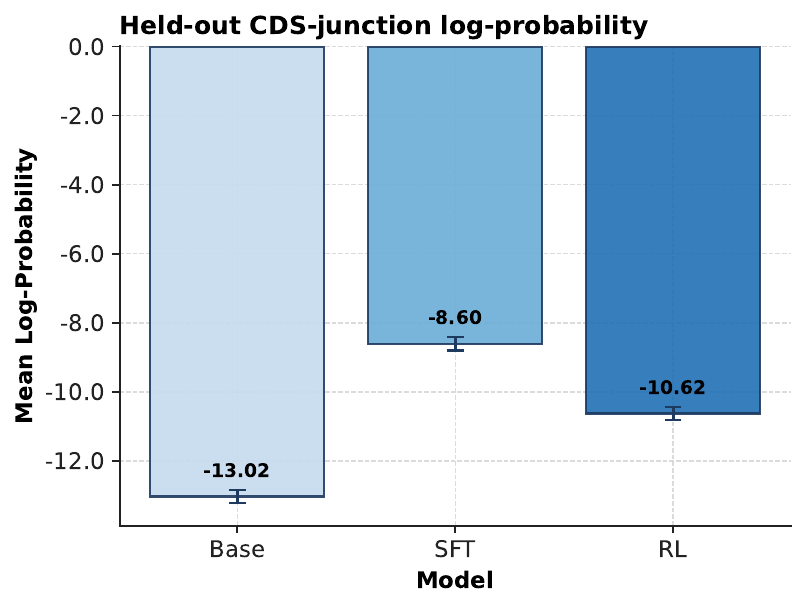}
\caption{CDS-junction log-probability on held-out plasmids. RL
achieves higher log-probability (lower surprisal) than Base despite
being trained on structural rewards only.}
\label{fig:surprisal}
\end{center}
\vskip -0.1in
\end{figure}

Across the 29 held-out plasmids, RL achieves higher CDS-junction
log-probability ($-10.62 \pm 0.19$ SEM) than Base ($-13.02 \pm 0.18$;
29/29 wins, Figure~\ref{fig:surprisal})---an improvement that persists
under per-nucleotide normalization ($-1.50 \pm 0.05$ vs.\
$-1.88 \pm 0.07$\,nats/nt; mean $\Delta = +0.37$\,nats/nt, 29/29 wins,
$p = 2.28 \times 10^{-13}$, $d = +2.41$)---suggesting that structural
reward shaping sharpens the model's distribution around real coding
boundaries rather than collapsing onto reward-only artifacts. As with
general continuations (\S\ref{sec:held_out}), SFT achieves the highest
CDS-junction log-probability on this set ($-8.60 \pm 0.20$ per token,
$-1.23 \pm 0.05$\,nats/nt; 29/29 vs.\ RL).

\section{Discussion}

\subsection{Effects of Structural Reward Shaping on Biophysical Properties}
\label{sec:discussion_shaping}

The central finding of this work is that a structurally motivated reward
function encoding functional annotations, length constraints, and
repeat penalties produces biophysical alignment with real plasmids
across properties at varying degrees of separation from the reward signal.
The ablation study (Table~\ref{tab:ablation}) demonstrates that the
cassette bonus, which rewards correct
promoter$\rightarrow$CDS$\rightarrow$terminator ordering, is the
critical component, and that individual reward terms are insufficient
in isolation.

The three-tier taxonomy introduced in \S\ref{sec:dist_comparison} shows that the biophysical
effects of RL post-training range from directly optimized properties
(length, component presence) through reward-correlated byproducts (GC
content) to indirectly shaped properties (3-mer composition) that arise
from the interaction of structural constraints
rather than any single reward term. Using the ablation values in
Table~\ref{tab:ablation}, MFE forms an approximately monotonic gradient
that tracks reward richness: the full reward yields the most stable
sequences ($-0.149$, matching the Addgene reference $-0.151$ in
Table~\ref{tab:ks_stats}), removing individual reward terms produces
intermediate MFE values ($-0.141$ to $-0.131$), and the
single-component variants (length-only $-0.126$, CDS-only $-0.103$)
sit closest to Base ($-0.105$). This suggests that real-plasmid-like
thermodynamic stability accumulates as multiple structural
constraints compose; no single reward term reproduces it in
isolation.

Held-out per-token continuation log-likelihood does not degrade relative to the
pretrained baseline on this benchmark (\S\ref{sec:held_out}): RL
outperforms Base on 29/29 non-Addgene plasmids for continuation and
29/29 for CDS-junction surprisal, with reduced cross-plasmid variance
(std $1.25 \to 1.09$). RL does underperform supervised fine-tuning on this same set, so we
do not claim that RL improves predictive performance over SFT or across
all plasmid domains. Within the engineered shuttle-vector scope of
the held-out set (\textit{Corynebacterium}, \textit{Mycobacterium},
\textit{Lactobacillus}, and SEVA backbones), RL's structural reward
shaping strictly improves per-token continuation over Base on all 29
plasmids (mean $\Delta = +0.83$ nats, 95\% CI $[+0.61, +1.05]$,
Cohen's $d = +1.41$; near-neutral per nucleotide, $\Delta = -0.02$\,nats/nt)
while improving CDS-junction likelihood both per token and per
nucleotide ($\Delta = +0.37$\,nats/nt, 29/29).

RL post-training has improved reasoning, instruction following, and
preference alignment in natural language processing, motivating its
application to biological sequence generation. The analogy has
important limits. In the common NLP post-training pipeline, RL often
follows supervised fine-tuning; here, RL starts directly from the
pretrained model and is evaluated as an alternative post-training path
alongside SFT to isolate structural reward shaping from supervised
corpus imitation. Evaluating sequential $\text{SFT}\rightarrow\text{RL}$
and measuring any alignment tax relative to an SFT initialization remain
open for future work. In NLP, reward
signals typically encode human judgments about text quality; here,
rewards encode structural constraints grounded in molecular biology.
Our results suggest that this grounding is advantageous: the reward
function selects for properties it does not explicitly encode (3-mer
composition, thermodynamic stability), consistent with the hypothesis that
structural constraints in DNA are tightly coupled to biophysical
outcomes---in part because selecting intact \textit{E.~coli} backbones and
Pyrodigal-scored ORFs brings their native local sequence statistics and
folding profiles. However,
biological sequence design ultimately requires function at the
organismal level: protein expression, host viability, and metabolic
compatibility are all properties that cannot be assessed from sequence
structure alone. This study evaluates only in silico structural properties.
Bridging the gap between structural validity and biological function
in a host organism remains the central open problem for RL-guided
biological design.

\subsection{Implications}

Structural reward shaping is a viable approach for constrained
biological sequence design. The five-model ablation and rejection
sampling baselines demonstrate that RL produces genuine distributional
shift: the base model's distribution contains very few valid plasmids
regardless of sample count, while RL concentrates probability mass on
functional regions of sequence space. This suggests that for generation
tasks with verifiable structural constraints, RL post-training offers
substantial advantages over supervised fine-tuning at low sampling
budgets and over naive sampling strategies. At very large sampling
budgets, rejection sampling can recover rare valid SFT outputs, but at
much higher sample cost.

\subsection{Limitations}

No generated sequences have been experimentally validated. While our QC
pipeline has been validated as a proxy for synthesis success in prior
work \cite{angus}, wet-lab validation remains the primary limitation and
is required before any practical deployment claims can be made.

RL optimization trades diversity for quality. pLannotate
\cite{plannotate} annotation of the 4{,}000 generated sequences shows
ColE1/pMB1/pBR322/pUC in 94.6\% (3{,}782/4{,}000) of sequences and p15A
in 12.9\% (515/4{,}000; Appendix~\ref{sec:diversity_appendix}), with a
dozen other origin classes appearing at $\leq$0.4\% each. Crucially,
restricting the audit to the 2{,}863 strict-QC-passing sequences shows
that 100\% carry an intact ColE1-family origin, whereas nearly all
p15A calls are partial fragments (410/414 QC-passing p15A hits occur
on the 500 rollouts seeded with the 100\,bp p15A prompt alongside a
generated ColE1 origin).
This pattern reflects the dominance of ColE1-based backbones in the
pretraining corpus: RL converges on the pretraining-dominant ColE1
distribution, as expected given that ColE1 is the most probable
valid replication origin in the Addgene-like engineered-plasmid
distribution that PlasmidGPT was trained on. The reward function
amplifies this existing bias rather than introducing it. Together with
strong within-prompt convergence on five of eight prompts
(Table~\ref{tab:per_prompt}), reduced diversity limits applications
requiring alternative copy-number regimes or host ranges; conditional
ORI prompting (Section~\ref{sec:future_work}) is the natural mitigation.

This study is scoped to engineered \textit{E.~coli} expression vectors
using Addgene as the target distribution. The reward function and QC
pipeline assume known functional annotations, so novel functional
elements (e.g., previously uncharacterized ORIs) will not be detected or
rewarded. Natural and environmental plasmid diversity (e.g., PLSDB) is
not addressed.

\subsection{Future Work}
\label{sec:future_work}

We plan to develop conditional generation capabilities where users can
specify desired plasmid properties (e.g., ``express protein X in
\textit{E.~coli} with high copy number''), which would both increase
practical utility and promote sample diversity through structured
prompting.

This work tests only GRPO; whether the same structural reward
landscape produces comparable results under alternative RL algorithms
(e.g., PPO, DPO, REINFORCE-style estimators with off-policy correction)
is an open question. A controlled comparison across post-training
algorithms on the same reward and base model would help separate
algorithm-specific behavior from properties of the reward itself.

\section{Conclusion}

We demonstrate that GRPO post-training with a structurally motivated
reward function achieves 71.6\% QC pass rate on whole-plasmid
generation compared to 4.3\% for the pretrained baseline, evaluated
across 8 prompts on 4{,}000 sequences. A five-model ablation
identifies the cassette bonus as the critical reward component, and
rejection sampling baselines confirm genuine distributional shift.
Beyond directly optimized properties, generated sequences align with
real engineered plasmids in 3-mer composition. These
effects can be understood through a three-tier
taxonomy distinguishing directly optimized, reward-correlated, and
indirectly shaped properties of reward-based optimization in biological
sequence space. Additionally, minimum free energy density converges to
the real-plasmid regime under both SFT and RL independently, despite
these being parallel post-training paths with different objectives.
Under the MFE metric tested here, structural reward shaping does not
appear to achieve high QC scores by producing physically unusual
sequence ensembles.

\section*{Acknowledgements}

We thank Linda Dekker for discussions and domain input that helped shape
the reward function design. McClain Thiel acknowledges support as an
Encode Fellow through Pillar VC's Encode Fellowship, part of Pillar
VC's Encode: AI for Science initiative. The Encode Fellowship is backed
by Pillar VC and powered by ARIA and DSIT.

\section*{Impact Statement}

This paper presents work whose goal is to advance the field of
computational biology and machine learning for biological sequence
design. The ability to generate novel, valid plasmid sequences could
accelerate research in synthetic biology, biomanufacturing, and
therapeutic development. While the immediate applications are primarily
beneficial for scientific research, we acknowledge the dual-use nature
of synthetic biology tools. The methods described here are limited to
generating plasmid sequences based on patterns in existing databases and
do not enable the design of harmful organisms without substantial
additional effort and expertise. We believe the benefits to research
efficiency and accessibility outweigh the potential risks, particularly
given existing biosafety regulations and oversight mechanisms in
synthetic biology research.

\newpage

\appendix

\section{Training Configuration Details}

\subsection{Supervised Fine-Tuning Hyperparameters}

\begin{itemize}
\item \textbf{Batch size:} 1
\item \textbf{Learning rate:} $5 \times 10^{-5}$ with 500 warmup steps
\item \textbf{Optimizer:} AdamW
\item \textbf{Epochs:} 3
\item \textbf{Gradient accumulation steps:} 8
\item \textbf{Hardware:} single NVIDIA L4 GPU
\end{itemize}

\subsection{Reinforcement Learning (GRPO) Hyperparameters}
\label{sec:rl_hparams}

Values below are Optuna-selected and live in the released
\texttt{plasmidrl/config.py} package module.

\begin{itemize}
\item \textbf{Per-device training batch size:} 16
\item \textbf{Generations per prompt (GRPO group size):} 4
\item \textbf{Policy learning rate:} $1.91 \times 10^{-5}$ (constant schedule)
\item \textbf{GRPO $\beta$ (KL coefficient):} $8.85 \times 10^{-4}$
\item \textbf{GRPO $\epsilon$ (clip range):} 0.265
\item \textbf{Loss type:} BNPO (Batch-Normalized Policy Optimization in HuggingFace \texttt{TRL}), reward scaling on
\item \textbf{Max prompt / completion tokens:} 1024 / 256
\item \textbf{Epochs:} up to 20; training terminated earlier on reward
plateau or collapse
\item \textbf{Training prompt types:} random 4--25\,bp seeds (excluding
ATG) and structured cassette seeds
\item \textbf{Hardware:} NVIDIA L40S (Anyscale)
\end{itemize}

\subsection{Reward Function Configuration}
\label{sec:reward_config}

The component weights below were chosen by subject-matter experts on
biological grounds before any RL training and were not tuned on the
evaluation set. The single Optuna-tuned reward parameter is the length
bonus scale (separate from the GRPO hyperparameters in
Appendix~\ref{sec:rl_hparams}, also
Optuna-selected); the count thresholds and component weights are fixed.
Source-of-truth values are in the released
\texttt{plasmidrl/ablations.py} and
\nolinkurl{plasmidrl/rewards/bioinformatics/scorer.py} package modules.

\begin{itemize}
\item \textbf{Origin of replication:} count exactly 1, weight = 1.0
\item \textbf{Selectable markers:} present/absent binary score, weight = 1.0
\item \textbf{Promoters:} count 1--5, weight = 1.0
\item \textbf{Terminators:} count 0--2, weight = 0.5
\item \textbf{Coding sequences:} count 1--2, weight = 1.0
\item \textbf{Promoter$\rightarrow$CDS$\rightarrow$terminator location bonus:}
location\_bonus\_scale = 0.5, proximity threshold 300\,bp, top-$K$ = 2
cassettes. Note that whenever $1 \leq n_{\text{cds}} \leq 2$, base
$\text{score}_{\text{cds}} = 1.0$ and the $[0,1]$ clip saturates;
however, on 5--7\,kb plasmids Pyrodigal routinely detects $n_{\text{cds}} > 2$
ORFs, zeroing base $\text{score}_{\text{cds}}$ so that the cassette
bonus (up to 0.5) supplies the primary positive CDS reward.
\item \textbf{Length prior:} length factor = 1 plus an extra bonus
inside the acceptable range; full extra bonus for sequences in
3--6\,kb; extra bonus tapers linearly to zero at the acceptable-range bounds
(2\,kb and 30\,kb); length factor = 0 outside [2\,kb, 30\,kb]. Bonus
scale = 0.7085 (Optuna-selected)
\item \textbf{Repeat penalty:} $-0.1$ per detected repeat region
$\geq$ 50\,bp (subject to the $[0, 1]$ reward clip)
\item \textbf{Count scoring and boundaries:} components below $\min_i$
receive $0.5 \times (n_i / \min_i)$ (which evaluates to 0 under the
configured $\min_i \in \{0,1\}$ thresholds); components above $\max_i$
receive 0 ($\text{violation\_penalty\_factor} = 0$). Setting
$\min_{\text{terminator}} = 0$ leaves standalone terminators optional
while rewarding cassette-associated terminators via
Algorithm~\ref{alg:cassette}. Rollout annotation, cassette proximity, and
canonical 50-mer repeat checks operate on linear generated strings.
\end{itemize}

\begin{algorithm}[H]
\caption{Plasmid Reward Scoring}
\label{alg:reward}
\begin{algorithmic}[1]
\STATE \textbf{Input:} DNA sequence $s$
\STATE \textbf{Output:} Reward score $R \in [0,1]$
\STATE
\STATE Annotate features using PlasmidKit (ORIs, promoters, terminators, markers) and Pyrodigal (CDS) $\rightarrow A(s)$
\STATE Merge overlapping same-type features (overlap threshold = 0.8); suppress CDS calls overlapping non-CDS features
\STATE
\FOR{each count-scored component $i \in \{$ori, promoter, terminator, cds$\}$}
\STATE Count features $n_i$ in $A(s)$
\IF{$n_i < \min_i$}
\STATE $\text{score}_i \leftarrow 0.5 \times (n_i / \min_i)$ if $\min_i > 0$ else $0.0$
\ELSIF{$n_i > \max_i$}
\STATE $\text{score}_i \leftarrow \text{violation\_penalty\_factor}$ \COMMENT{$= 0$; violating count zeroes the component}
\ELSE
\STATE $\text{score}_i \leftarrow 1.0$
\ENDIF
\ENDFOR
\STATE $\text{score}_{\text{cds}} \leftarrow \text{CassetteBonus}(A(s), \text{score}_{\text{cds}})$ \COMMENT{Algorithm~\ref{alg:cassette}}
\STATE $\text{score}_{\text{marker}} \leftarrow 1.0$ if at least one marker is present, else $0.0$
\STATE
\STATE \textbf{Compute normalized weighted base score and length factor:}
\STATE $I \leftarrow \{\text{ori, promoter, terminator, marker, cds}\}$ \COMMENT{$\sum_{i \in I} w_i = 4.5$}
\STATE $\text{base} \leftarrow \left(\sum_{i \in I} w_i \times \text{score}_i\right) / \left(\sum_{i \in I} w_i\right)$
\STATE $\text{length\_factor} \leftarrow$ length score for $s$
\STATE $\text{reward} \leftarrow \text{clip}(\text{base} \times \text{length\_factor}, 0, 1)$
\STATE
\STATE \textbf{Detect repeat regions (linear string):}
\STATE Find canonical $k$-mers ($k=50$) appearing $\geq 2$ times
\STATE Merge overlapping occurrences $\rightarrow \mathcal{R}$ regions
\STATE $\text{repeat\_penalty} \leftarrow |\mathcal{R}| \times 0.1$
\STATE
\STATE \textbf{return}\ $R \leftarrow \text{clip}(\text{reward} - \text{repeat\_penalty}, 0, 1)$
\end{algorithmic}
\end{algorithm}

\begin{algorithm}[H]
\caption{CDS Cassette Bonus Subroutine}
\label{alg:cassette}
\begin{algorithmic}[1]
\STATE \textbf{Input:} Annotated features $A(s)$, base $\text{score}_{\text{cds}}$
\STATE \textbf{Output:} Updated $\text{score}_{\text{cds}}$
\STATE
\FOR{each promoter $p$ in $A(s)$}
\STATE Find same-strand CDS $c$ minimizing edge-to-edge gap $d(p,c)$ ($0$ if overlapping)
\STATE Find same-strand terminator $t$ minimizing $d(c,t)$
\STATE $\text{pts} \leftarrow 0$; award for $(p,c)$ and (if $t$ exists) $(c,t)$:
\STATE \hspace{0.5cm} upstream$\rightarrow$downstream strand order: $+5$ pts
\STATE \hspace{0.5cm} gap $\leq 300$\,bp (if ordered / if out-of-order): $+5$ / $+3$ pts
\ENDFOR
\STATE Sum top $K=2$ cassettes ($\max 20$ pts each) $\rightarrow \text{cassette\_points}$
\STATE $\text{cassette\_bonus} \leftarrow \text{location\_bonus\_scale} \times (\text{cassette\_points} / (20 K))$ \COMMENT{$\text{location\_bonus\_scale}=0.5$}
\STATE \textbf{return}\ $\text{score}_{\text{cds}} \leftarrow \text{clip}(\text{score}_{\text{cds}} + \text{cassette\_bonus}, 0, 1)$
\end{algorithmic}
\end{algorithm}

\subsection{Sampling Parameters}
\label{sec:sampling_params}

Evaluation temperatures differ between protocols, held constant within
each. Top-$p$ is 0.90 and repetition penalty is 1.0 throughout, and
generations are capped at 256 BPE tokens ($\approx$ 5--15\,kb of DNA).
Table~\ref{tab:sampling_params} summarizes the temperature settings
used for each evaluation.

\begin{table}[H]
\caption{Sampling temperature per evaluation protocol (reward ablations
use $0.95$ in Table~\ref{tab:ablation} and $1.15$ in
Fig.~\ref{fig:qc_breakdown}; rejection sampling evaluates RL at matched
$1.00$ and sweep-optimal $1.15$).}
\label{tab:sampling_params}
\vskip 0.15in
\begin{center}
\begin{small}
\begin{sc}
\resizebox{\columnwidth}{!}{%
\begin{tabular}{lc}
\toprule
Protocol & $T$ \\
\midrule
Main 8-prompt evaluation (Tables~\ref{tab:qc_results},~\ref{tab:novelty}) & 1.00 \\
Reward ablation (Appendix~\ref{sec:ablation_appendix}) & 0.95, 1.15 \\
Rejection sampling top-$K$ (Table~\ref{tab:baselines}), Base & 1.00 \\
Rejection sampling top-$K$ (Table~\ref{tab:baselines}), SFT & 1.00 \\
Rejection sampling top-$K$ (Table~\ref{tab:baselines}), RL & 1.00, 1.15 \\
\bottomrule
\end{tabular}
}
\end{sc}
\end{small}
\end{center}
\vskip -0.1in
\end{table}

\subsection{Distribution Comparison Metrics}

Table~\ref{tab:ks_stats} presents summary statistics
comparing the distributions of generated plasmids to real plasmids
across key biophysical metrics. For scalar summaries, closeness to the
Real column indicates alignment; for 3-mer JSD, we compute the base-2
plasmid-to-centroid divergence
$\mathrm{JSD}_2(p_s \| p_{\text{ref}}) = \frac{1}{2} D_{\mathrm{KL}}(p_s \| m) + \frac{1}{2} D_{\mathrm{KL}}(p_{\text{ref}} \| m)$
in bits, where $p_s$ is the 64-bin 3-mer distribution of sequence $s$,
$p_{\text{ref}}$ is the pooled 3-mer distribution of the 500-plasmid
Addgene reference panel, and $m = \frac{1}{2}(p_s + p_{\text{ref}})$
(pairwise Real-vs-Real JSD averages $0.012$\,bits). To control for
sequence length in 3-mer histogram estimation, a sensitivity check over
fixed non-overlapping $1{,}000$\,bp windows across all sequences
$\geq 1$\,kb yields comparable means across groups (Real $0.031$,
$n{=}500$; Base $0.030$, $n{=}1{,}465/4{,}000$; SFT $0.030$,
$n{=}3{,}999$; RL $0.029$, $n{=}3{,}838$), confirming that Base's
elevated whole-sequence JSD ($0.171$) stems from its $63.4$\% short
($<1$\,kb) truncated outputs and finite-sample histogram variance.

\begin{table}[H]
\caption{Distribution comparison metrics (Addgene reference, $n{=}500$;
all models on 4{,}000 samples, 8-prompt protocol; MFE: Mathews 2004
DNA parameters).}
\label{tab:ks_stats}
\vskip 0.15in
\begin{center}
\begin{small}
\begin{sc}
\resizebox{0.95\columnwidth}{!}{%
\begin{tabular}{lcccc}
\toprule
Metric & Real & Base & SFT & RL \\
\midrule
Mean Length (bp) & 7{,}469 & 1{,}949 & 5{,}441 & 6{,}517 \\
GC Content & 0.510 & 0.484 & 0.493 & 0.524 \\
3-mer JSD & 0.006 & 0.171 & 0.016 & 0.022 \\
MFE Density & $-0.151$ & $-0.105$ & $-0.148$ & $-0.149$ \\
\bottomrule
\end{tabular}
}
\end{sc}
\end{small}
\end{center}
\vskip -0.1in
\end{table}

\subsection{Reference Plasmids for Distribution Comparison}
\label{sec:reference_plasmids}

For the distribution comparison in Section~\ref{sec:dist_comparison}, we use a curated set of
500 engineered plasmids drawn from the Addgene repository.
These plasmids span common backbones, selection markers, and use-cases
for protein expression and genome editing, and serve as a reference
panel for evaluating how closely model-generated plasmids match real
engineered constructs.

This same 500-plasmid Addgene reference panel---spanning engineered
vectors across cloning, protein expression, and genome editing without
host-species restriction---also provides the real-plasmid 21-mer
Jaccard diversity reference ($0.848$) in Table~\ref{tab:qc_results}.

All minimum free energy (MFE) calculations use ViennaRNA 2.7.2 with
Mathews 2004 DNA energy parameters \cite{viennarna}. In the main
evaluation, sequences $<$50\,bp are excluded (assigned $0.0$ in raw Base
summaries), sequences of 50--3{,}000\,bp are folded globally as circular
DNA (\texttt{circ=1}), and sequences $>$3{,}000\,bp use averaged linear
500\,bp sliding windows (\texttt{circ=0}, 250\,bp stride). Applying the
uniform linear 500\,bp sliding-window protocol (\texttt{circ=0}, 250\,bp
stride) across all sequences $\geq$500\,bp yields mean MFE densities
($\pm$SEM) of $-0.150 \pm 0.0006$ for Real ($n{=}500/500$),
$-0.145 \pm 0.0004$ for SFT ($n{=}3{,}999/4{,}000$),
$-0.154 \pm 0.0003$ for RL ($n{=}3{,}838/4{,}000$), and
$-0.143 \pm 0.0013$ for Base ($n{=}1{,}486/4{,}000$), confirming that
Base's all-sequence $-0.105$ is driven by its $2{,}514$ sequences
$<$500\,bp rather than the 3\,kb folding-mode boundary.

\section{Reward Ablation Study}
\label{sec:ablation_appendix}

\begin{figure}[H]
\vskip 0.15in
\begin{center}
\includegraphics[width=\columnwidth]{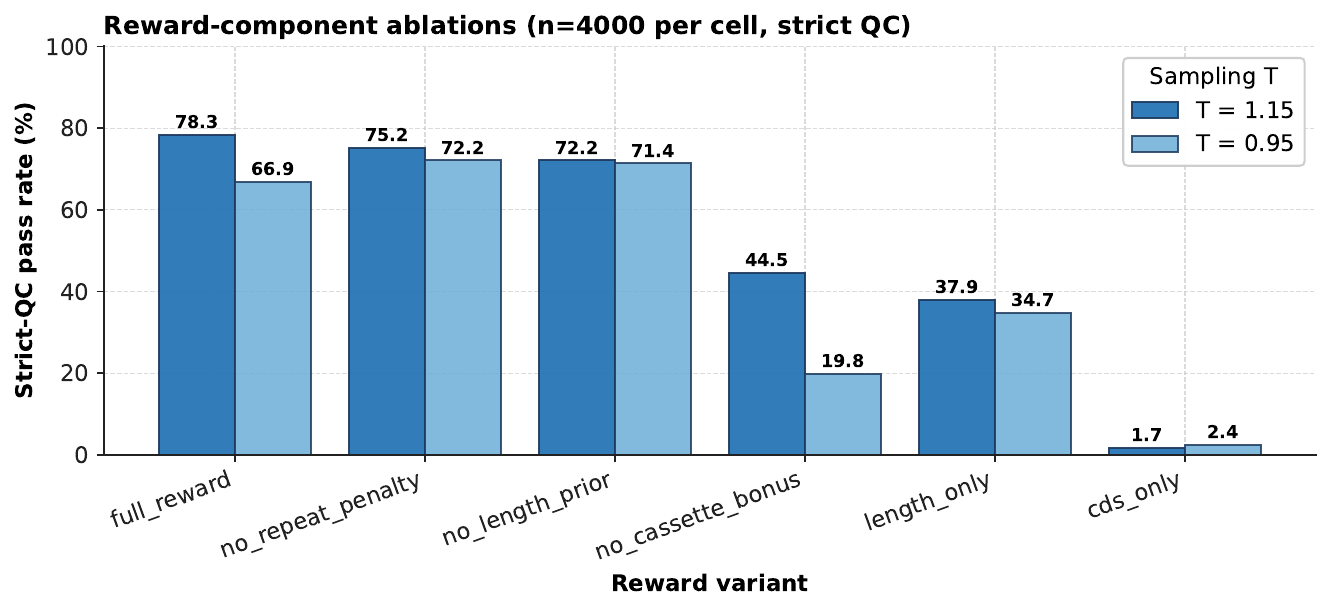}
\caption{QC pass rates across reward ablation variants. Light bars
show $T{=}0.95$ (matching Table~\ref{tab:ablation}); dark bars show a
$T{=}1.15$ sensitivity comparison. Removing the cassette bonus
produces the largest drop at both temperatures, identifying it as
the critical reward component.}
\label{fig:qc_breakdown}
\end{center}
\vskip -0.1in
\end{figure}

To identify which reward components drive the observed improvements, we
train five ablation models by selectively disabling individual reward
terms. Each model is evaluated on 4{,}000 sequences using the same
8-prompt protocol as the main evaluation. Table~\ref{tab:ablation}
presents the results and Figure~\ref{fig:qc_breakdown} visualizes the
pass-rate column.

\begin{table}[H]
\caption{Reward ablation results (all rows at $T{=}0.95$;
$n{=}4{,}000$ per model). Len.\ is mean cleaned sequence length (bp); MFE
is mean per-base-pair minimum free energy density (kcal/mol/bp, Mathews
2004 DNA parameters); Div.\ is mean pairwise 21-mer Jaccard distance.}
\label{tab:ablation}
\vskip 0.15in
\begin{center}
\begin{small}
\begin{sc}
\resizebox{\columnwidth}{!}{%
\begin{tabular}{lcccc}
\toprule
Model & Pass & Len.\ (bp) & MFE & Div. \\
\midrule
RL (full reward)    & 66.9\% & 6{,}495 & $-0.149$ & 0.578 \\
No repeat penalty   & 72.2\% & 6{,}563 & $-0.141$ & 0.394 \\
No length prior     & 71.4\% & 5{,}460 & $-0.131$ & 0.447 \\
No cassette bonus   & 19.8\% & 6{,}184 & $-0.134$ & 0.355 \\
Length only         & 34.7\% & 3{,}946 & $-0.126$ & 0.760 \\
CDS only            & 2.4\%  & 2{,}012 & $-0.103$ & 0.823 \\
\bottomrule
\end{tabular}
}
\end{sc}
\end{small}
\end{center}
\vskip -0.1in
\end{table}

The cassette bonus, which rewards correct
promoter$\rightarrow$CDS$\rightarrow$terminator ordering and proximity,
is the single most important reward component. Removing it drops pass
rate from 66.9\% to 19.8\% at $T{=}0.95$ (and from 78.3\% to 44.5\% at
$T{=}1.15$) while collapsing diversity to 0.355, the lowest of any
ablation. At $T{=}0.95$, removing the repeat penalty (72.2\%) or length
prior (71.4\%) yields a higher raw pass rate than the full reward
(66.9\%) while sharply reducing diversity (0.394 and 0.447 vs.\ 0.578)
and weakening MFE stability ($-0.141$ and $-0.131$ vs.\ $-0.149$),
reflecting convergence onto lower-diversity, repeat-free canonical
backbones (note also that \texttt{No length prior} retains a 5{,}460\,bp
mean length, with 88.3\% of sequences $>3$\,kb). At $T{=}1.15$
(Figure~\ref{fig:qc_breakdown}), this ranking reverses and the full
reward outperforms all ablations (78.3\% vs.\ 75.2\% and 72.2\%).
Training with only the CDS reward (2.4\%) or length reward alone
(34.7\%) yields substantially lower pass rates, confirming that the full
reward function's components act synergistically. Note that
Figure~\ref{fig:qc_breakdown}'s $T{=}1.15$ full-dataset pass rate for
\texttt{full\_reward} (78.3\%, $n{=}4{,}000$) differs slightly from
Table~\ref{tab:baselines}'s $K{=}1$ bootstrap estimate (76.8\%,
$M{\times}8{=}400$ draws from a 10{,}000-sequence pool).

The full-reward row uses the same RL model as
Table~\ref{tab:qc_results} and Table~\ref{tab:ks_stats}, sampled at
$T{=}0.95$ (the original ablation setpoint from the hyperparameter sweep
configuration) rather than $T{=}1.0$. Pass rate and diversity differ
modestly with sampling temperature ($66.9\%$ vs.\ $71.6\%$; $0.578$
vs.\ $0.596$); MFE is essentially unchanged ($-0.149$ in both
tables).

MFE values across ablations are discussed in
\S\ref{sec:discussion_shaping}.

Functional diversity (origin-of-replication families) is reported
separately in Appendix~\ref{sec:diversity_appendix} from a pLannotate
audit over the full 4{,}000-sequence generations.

\section{Functional Diversity of Generated Sequences}
\label{sec:diversity_appendix}

The 21-mer MinHash Jaccard diversity reported in the main text
(0.596 for the full RL model) measures sequence-level variation,
which on a fixed-backbone plasmid is dominated by cargo and cassette
arrangement rather than architectural diversity. A more direct measure
of architectural diversity is the distribution over origin-of-replication
(ORI) types: the ORI is the strongest determinant of copy number, host
range, and compatibility, and is what a downstream user is typically
choosing between.

To audit how much functional diversity survived RL training, we
annotated all 4{,}000 GRPO-at-$T{=}1.0$ generations with
pLannotate \cite{plannotate}, separately from the validity QC pipeline
used in Section~\ref{sec:results}, and cross-tabulated the hits on the
2{,}863 strict-QC-passing subset. Table~\ref{tab:ori_breakdown} reports
the resulting ORI-family counts.

\begin{table}[H]
\caption{pLannotate ORI breakdown on all 4{,}000 RL sequences ($T{=}1.0$)
and the 2{,}863 strict-QC-passing subset. Counts are unique sequences per
origin family. Asterisks ($^*$) indicate families whose QC-passing calls
are partial fragments ($\geq$80\% \texttt{fragment=True} in pLannotate)
co-occurring with a full-length ColE1 origin.}
\label{tab:ori_breakdown}
\vskip 0.15in
\begin{center}
\begin{small}
\begin{sc}
\resizebox{\columnwidth}{!}{%
\begin{tabular}{lrr}
\toprule
Origin family & All Seq.\ ($n{=}4{,}000$) & QC Pass ($n{=}2{,}863$) \\
\midrule
ColE1 / pMB1 / pBR322 / pUC          & 3{,}782 & 2{,}863 \\
p15A                                 &    515 & 414$^*$ \\
incP                                 &     16 &  13$^*$ \\
Yeast 2$\mu$                         &     15 &   9$^*$ \\
EBV oriP (variants)                  &     15 &   9$^*$ \\
RSF1030                              &     12 &   2$^*$ \\
pVS1 ({\normalfont\itshape Pseudomonas}) &      5 &   5$^*$ \\
{\normalfont\itshape S.~cerevisiae} ARS1 &      5 &   4$^*$ \\
M13 / pSC101 / SV40                  &  $\leq$4 each & $\leq$3$^*$ each \\
F plasmid                            &      1 &   1$^*$ \\
\bottomrule
\end{tabular}
}
\end{sc}
\end{small}
\end{center}
\vskip -0.1in
\end{table}

3{,}797 of the 4{,}000 sequences carry at least one detectable origin
signature (3{,}795 excluding f1 bacteriophage origins). Across all
generations, the two well-represented \textit{E.~coli} replication
origins are ColE1 (in 94.6\% of sequences; 3{,}782 unique) and p15A
(in 12.9\%; 515 unique); a dozen further origin classes are detected
occasionally in trace quantities. We do not count the f1 bacteriophage
origin (2{,}842 unique sequences; 2{,}859 hits) toward
replication-strategy diversity because it co-occurs with ColE1 in
pUC-style backbones for ssDNA production rather than serving as an
independent replication origin.

Crucially, restricting the audit to the 2{,}863 strict-QC-passing
sequences resolves the apparent multi-ORI overlap: 100\%
(2{,}863/2{,}863) of QC-passing sequences carry a full-length
ColE1-family origin, whereas all non-ColE1 pLannotate hits among QC
passers are partial fragment matches (98.6\% of p15A calls and 100\% of
incP, Yeast 2$\mu$, EBV oriP, RSF1030, and ARS1 calls carry
\texttt{fragment=True}). In particular, 410 of the 414 QC-passing p15A
detections (and 500 of 515 overall) occur on the 500 rollouts seeded
with the 100\,bp \texttt{pACYC184 (p15A)} prompt alongside a generated
ColE1 origin; de novo p15A generation on unseeded prompts is rare
(0.43\%, 15/3{,}500 overall; 0.16\%, 4/2{,}453 among non-p15A-seeded
QC passers). Because strict BLAST QC requires $\geq$99\%
coverage against OriDB, it registers only the intact ColE1 origin.

This pattern reflects the dominance of ColE1-based backbones in the
pretraining corpus (predominantly ColE1-based \textit{E.~coli}
expression vectors, with p15A as the second most common backbone
family); RL converges on this distribution because it is already the
most probable regime in the pretrained model. RL narrows but does not
collapse the distribution over origins.
Reduced architectural diversity remains a real limitation for
applications that need alternative copy-number regimes or host
ranges; conditional generation that exposes the ORI as a control
input (Section~\ref{sec:future_work}) is the planned mitigation.

\section{Per-Prompt QC Pass Rates}
\label{sec:per_prompt_appendix}

Table~\ref{tab:per_prompt} breaks down Base, SFT, and RL QC pass rates
alongside RL within-prompt sequence-level diversity across the eight
evaluation prompts.

\begin{table}[H]
\caption{Per-prompt strict-QC pass rates (Base, SFT, RL) and
within-prompt 21-mer Jaccard diversity (RL) at $T{=}1.0$ (500 sequences
per prompt).}
\label{tab:per_prompt}
\vskip 0.15in
\begin{center}
\begin{small}
\begin{sc}
\resizebox{\columnwidth}{!}{%
\begin{tabular}{lrcccc}
\toprule
Prompt Type & bp & Base & SFT & RL & RL Div.\\
\midrule
Start codon (ATG)         & 3   & 1.0\%  & 13.6\% & 88.4\% & 0.13 \\
Random seed               & 10  & 6.0\%  & 14.4\% & 85.0\% & 0.13 \\
Random seed               & 25  & 1.0\%  & 15.4\% & 71.0\% & 0.10 \\
pACYC184 (p15A) ORI       & 100 & 7.4\%  & 11.0\% & 82.0\% & 0.10 \\
pUC19 ORI region          & 100 & 0.0\%  & 3.6\%  & 14.4\% & 0.81 \\
KanR cassette             & 300 & 12.2\% & 7.2\%  & 80.2\% & 0.71 \\
CMV enhancer              & 300 & 0.0\%  & 15.8\% & 96.6\% & 0.86 \\
GFP cassette              & 917 & 6.6\%  & 6.8\%  & 55.0\% & 0.08 \\
\midrule
Overall / Mean            & --- & 4.3\%  & 11.0\% & 71.6\% & 0.365 \\
\bottomrule
\end{tabular}
}
\end{sc}
\end{small}
\end{center}
\vskip -0.1in
\end{table}

\end{document}